\pdfoutput=1

\documentclass{article}
\usepackage{stywhispers,amsmath,epsfig}

\title{Understand customer reviews with less data and in short time: pretrained language representation and active learning}
\name{Yanwei CUI, ~~ Xavier ILLY}
\address{GMF Assurances, Groupe Cov\'ea\\
148 Rue Anatole France, 92300 Levallois-Perret, France\\
ycui@gmf.fr, ~ xavier.illy@covea.fr}
\begin{document}
%
\maketitle
\begin{abstract}
In this paper, we address customer review understanding problems by using supervised machine learning approaches, in order to achieve a fully automatic review aspects categorisation and sentiment analysis. In general, such supervised learning algorithms require domain-specific expert knowledge for generating high quality labeled training data, and the cost of labeling can be very high. To achieve an in-production customer review machine learning enabled analysis tool with only a limited amount of data and within a reasonable training data collection time, we propose to use pre-trained language representation to boost model performance and active learning framework for accelerating the iterative training process. The results show that with integration of both components, the fully automatic review analysis can be achieved at a much faster pace. 

\end{abstract}
\section{Introduction}

Natural language processing has gain continuously attention in recent years, not only for academe research purposes but also for a real-world use case in various industrial sectors. Advanced neural architectures achieve significant improvements on difficult language understanding problems, thus enable various applications such as  named-entity recognition \cite{lample2016neural}, semantic role labeling \cite{he2017deep}, sentiment analysis or opinion mining \cite{zhang2018deep, wang2016attention}, machine translation \cite{wu2016google}, etc. 

Thanks to the recent advances in the language understanding fields with the help of deep learning, a large number of machine learning projects turn from academic research outcomes into industrial products. For instance, the neural machine translation system \cite{wu2016google} now delivers a very high-quality translation that is approaching human-level accuracy; well trained neural models have been proposed to business users as prediction services on the cloud to perform difficult tasks such as topic extraction, sentiment analysis. These advances in technologies have enabled the need for automatic language analysis in the marketing, financial institutes, and others.

However, despite typical applications that can share the trained model to perform universal tasks such as speech-to-text, translation, most of the applications often require to train a custom model with company-owned data and strongly rely on domain-specific knowledge. For example, in an assurance company, one might interest in investigating customer comments on the topic related to processing time of insurance claims, while for the e-commercial website, reviews about the quality of the goods are more interesting to dive deeper. It might be challenging to analyse these problems without importing the specificities of their data and train a custom model based on these data. 

In addition, most of the advanced neural network architectures require a considerable amount of labeled data for training, and labeling is often tedious and time-consuming. The challenges of labeling data can significantly slow down the development of machine learning enabled projects for companies.

In this paper, we address the customer review understanding problems with deep learning framework, with a particular focus on two methods that can accelerate the training by adopting the pretrained model and active learning (in Sec.\;\ref{method}). The preliminary results (in Sec.\;\ref{exper}) show that the iterative process of training robust neural network models can be significantly shorted, thus saving a large amount of cost. We also provide a conclusion and further directions in the end.

\section{Model architecture}
\label{method}

\subsection{Recurrent neural network with pretrained embedding}

Recurrent neural networks have been widely used for sequence formed data, as the capability of taking the input as various lengths and learning the dependency among elements at different positions. The applications of recurrent neural networks cover from time series analysis \cite{cui2018modelling}, speech recognition \cite{lin2019lstm}, to natural language processing \cite{wang2016attention}, and continuous gain attention in different research domains.

In the field of natural language processing, one key component is embedding or so-called language representation. The objective is to pass the words into a fixed-length vector, which could be later processed and fed into the classifiers. A straightforward idea is one-hot encoding, which represents each word in a sparse dictionary with one indicating its position among all possible words. The distribution of all the words inside a sentence can be directly used to represent that sentence.  Such an approach often refers as bag-of-words \cite{mikolov2013efficient}. Some improvements have been added to such architecture by training a matrix projection over the words \cite{joulin2016bag}. Recent successes are focused on embedding with pretrained language model \cite{devlin2018bert}, which allows a semantic representation of words or tokens in the sentences, and the aggregated of embedded vectors often leads to a better result compared to word distribution with one-hot encoding.  

\begin{figure}[t]
	\centering
	\includegraphics[width=0.48\textwidth]{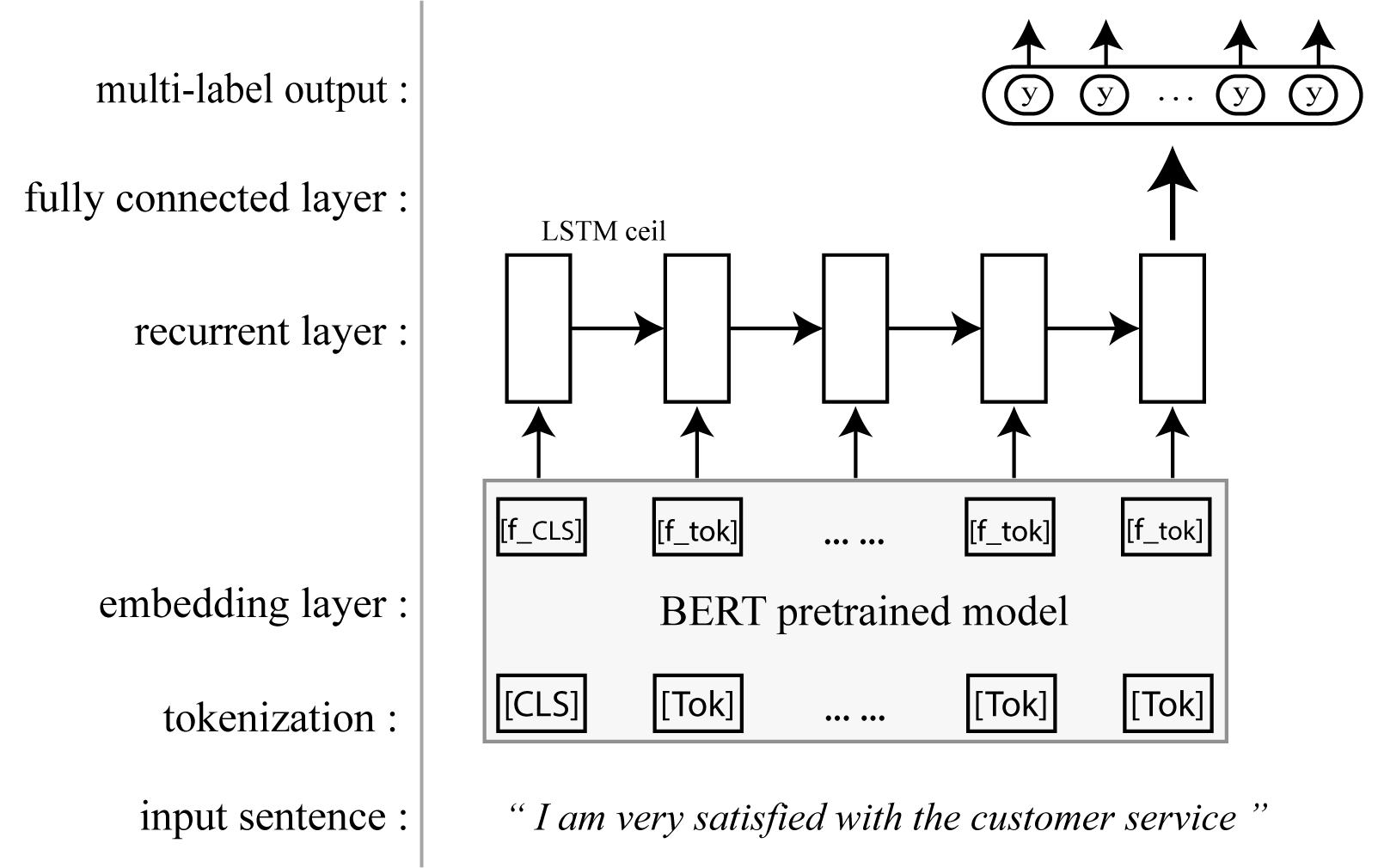}
	\caption{The architecture of our recurrent neural network model.}
	\label{fig:model}
\end{figure}

The pretrained models, such as BERT\cite{devlin2018bert} and ELMo \cite{peters2018deep}, have shown very promising results on various tasks. With the access of unlimited texts written on the Internet, these models can capture the semantic meaning of the language without being dedicated to specific tasks. The results of the pretrained model can later be used as preliminary inputs for various tasks, including classification, named entity recognition, question answering, language common-sense inference. 

In our architecture, as illustrated in Fig.\;\ref{fig:model}, we use BERT \cite{devlin2018bert} as our pretrained embedding module. The outputs of BERT embedding on the tokens (words) are then fed into a recurrent neural network. Long short-term memory (LSTM) \cite{hochreiter1997long}, as we used here for taking into account of the long dependencies among words. The output of the last LSTM ceil is coupled with fully connected layers, where the final output as the sigmoid active function that allows multi-label outputs. Unlike softmax layer that normalizes the output into probability assigned to each class (commonly used for single label classification problem), the sigmoid activated layer output unnormalized probability of $[0,1]$ for each class, with all the output values close to $1$ indicating a high probability of the underlying sentences belongs to these classes. The multi-label loss function of our recurrent network is computed as the sum of binary cross-entropy between the prediction and true label for each class: 
\begin{equation}
\text{\textit{Multi-label loss}} =  \sum_{i = c_1}^{C_N} y_i \log(\hat{y}_i) + (1- y_i) \log(1 - \hat{y}_i )
\end{equation}

\subsection{Active learning strategy}

In the supervised learning framework, a large amount of labeled data are often required to archive excellent performance, and it is especially true for training a neural network. However, labeling data can be time-consuming and increase the cost of machine learning projects. 

\begin{figure}[htb]
	\centering
	\includegraphics[width=0.5\textwidth]{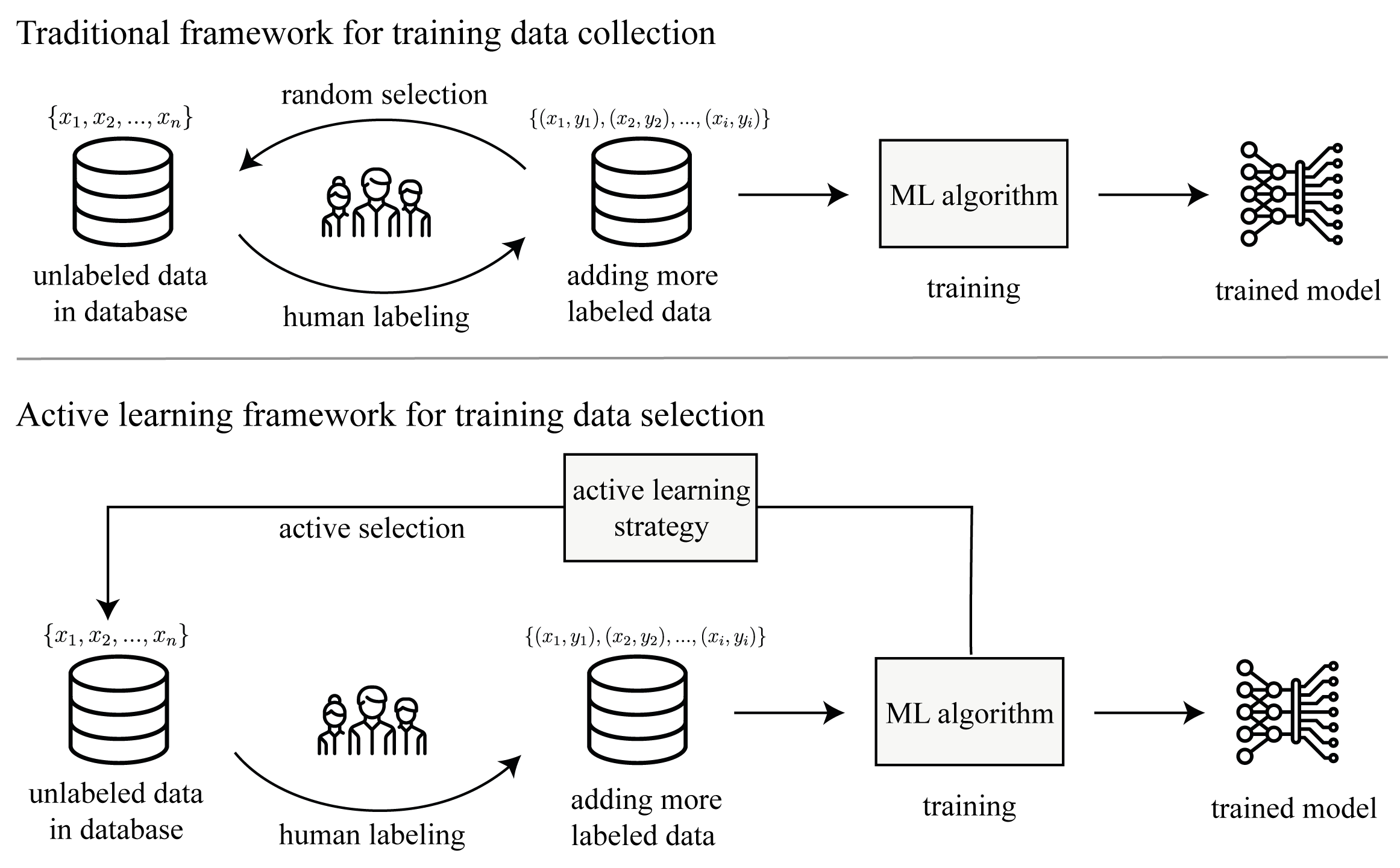}
	\caption{Traditional data collection framework \textit{v.s.} active learning framework.}
	\label{fig:active}
\end{figure}

In the conventional data collection process (as illustrated in Fig.\;\ref{fig:active}), human labeling tasks are conducted in a random form. The experts use their domain knowledge to label the data in the database by randomly selecting the data samples, and provide the labeled data for training. To further improve the performance, it often requires a more considerable amount of labeled data, and it is often conducted by other batches of a random selection of the data to label. Such an iterative process is highly insufficient as the continuous learning process is cutting into two different groups of subtasks without communication between them.

In contrast to the conventional data selection method,  active learning offers an alternative strategy for collecting the supervised training data, as illustrated in Fig.\ref{fig:active}. The active learning strategies choose the samples that need to be labeled, with the aim of maximizing the machine learning algorithm's performance \textit{w.r.t} each incremental labeled dataset. These strategies include: Least Confidence \cite{culotta2005reducing}, Bayesian Active Learning by Disagreement \cite{gal2016theoretically},  core-set selection \cite{sener2017active}, etc. where all proposed strategies can be defined within a common framework: train the model with existing labeled data,  use the trained model to select (under proposed measurement) the candidate from a pool of unlabeled dataset, label the selected candidate data points, and train the new model with augmented training dataset, as illustrated in Fig.\;\ref{fig:active}.

In this paper, in order to show the effectiveness of adopting the active learning framework, we use one straightforward uncertainty-based strategy in the multi-label classification cases. The uncertainty score is measure by:

\begin{equation}
\text{\textit{Uncertainty score}}  = 1 - \max_{i}(\hat{y}_i) 
\label{eq:certain}
\end{equation}

where we choose the unlabeled data with the lowest predicted probabilities among all classes. Intuitively speaking, the model can improve itself by seeing more diverse samples that are not semantical similar to the training set or model cannot confidently predict its labels. In such a setting, the model selects the data instances that are difficult to be assigned to any classes (uncertainty) for interactive labeling with human experts.

\section{Experiments and discussions}
\label{exper}

We evaluate the proposed method by using a real-world customer review dataset, with 6929 instances for training and 1456 instances for evaluation. The evaluations are conducted on two separate multi-label classification tasks: aspects categorisation (13 classes) and sentiment analysis (2 classes), as shown in Tab.\;\ref{tab:data}.  

\begin{table}[!htb]
	\centering
	\small
	\caption{\small Number of collected data for training and validation for multi-label classification: aspect categorisation and sentiment analysis.}
	
	\begin{tabular}{l|c|c}
		
		\hline
		\textbf{Aspect category} &  \textbf{ training} & \textbf{  validation} \\
		
		Internet usage & 330 & 67 \\
		Global Management  & 2562 & 525 \\
		Loyalty & 1078 & 203 \\
		Contract  & 347 & 61 \\
		Financial  & 776 & 184 \\
		Accessibility &  730 & 129 \\
		Reception & 959 & 237 \\
		Empathy  & 1144 & 184 \\
		Information provided & 1184 & 215 \\
		Processing time  & 1845 & 379 \\
		Visibility  & 603 & 147 \\
		Expert  & 442 & 92 \\
		Repairing & 427 & 94 \\
		\hline
		\hline
		\textbf{Sentiment polarity} & & \\
		
		Positive & 8254 & 1664 \\
		Negative & 4173 & 853 \\
		\hline
		\hline
		\textbf{Total instances} & 6929 & 1456 
	\end{tabular}
\label{tab:data}
\end{table}

We evaluate three different settings: 1. CNN embedding \cite{cui2018modelling,zhang2015character} with random samples 2. BERT pretrained embedding with random samples 3. BERT pretrained embedding with active learning selected samples. 

All three settings are connected to a recurrent neural network (two layer of LSTM) and a fully connected layer with sigmoid active function for multi-label outputs, implemented with TensorFlow library \cite{45381} in Python. 

\begin{figure}[htb]
	\centering
	\includegraphics[width=0.46\textwidth]{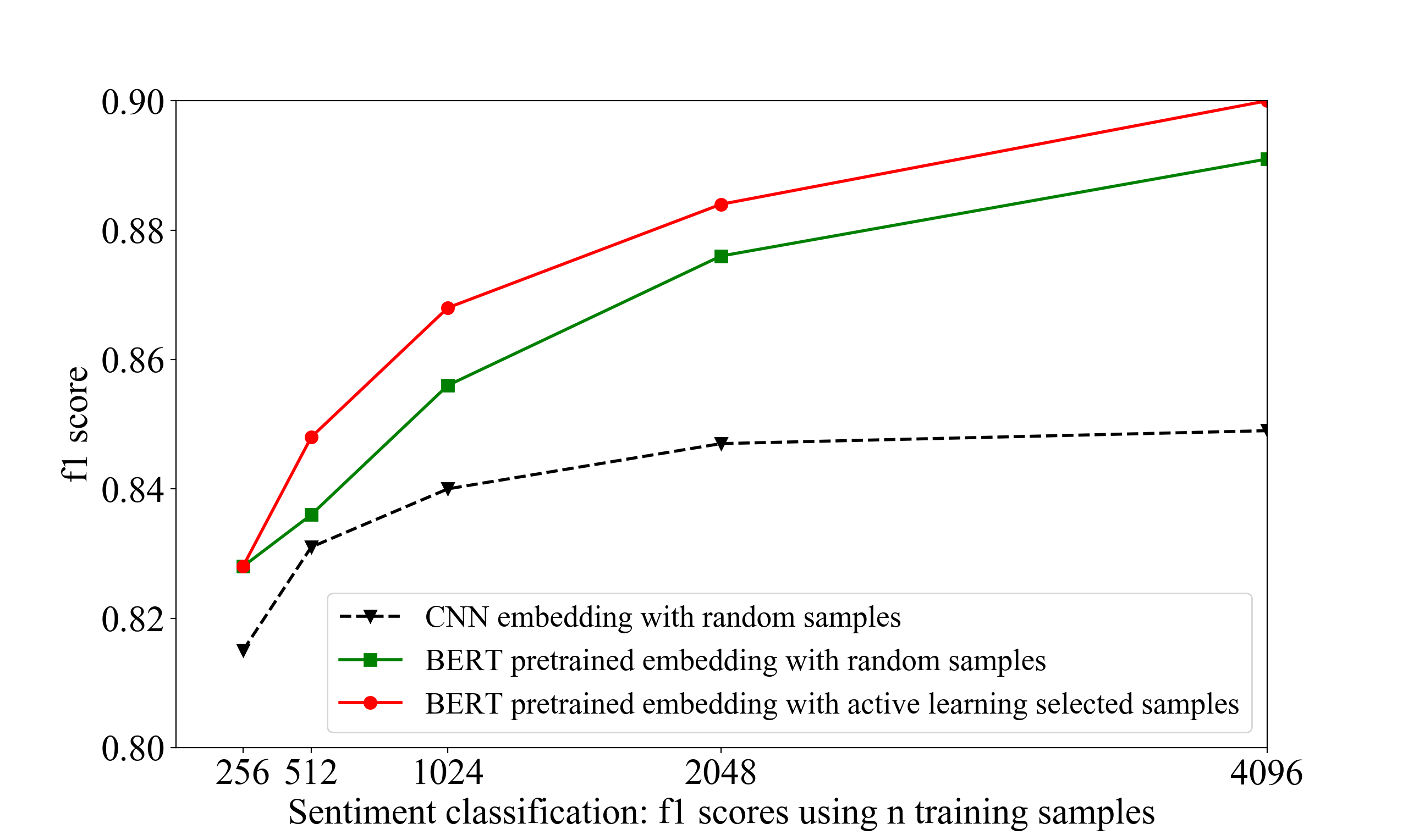}
	\includegraphics[width=0.46\textwidth]{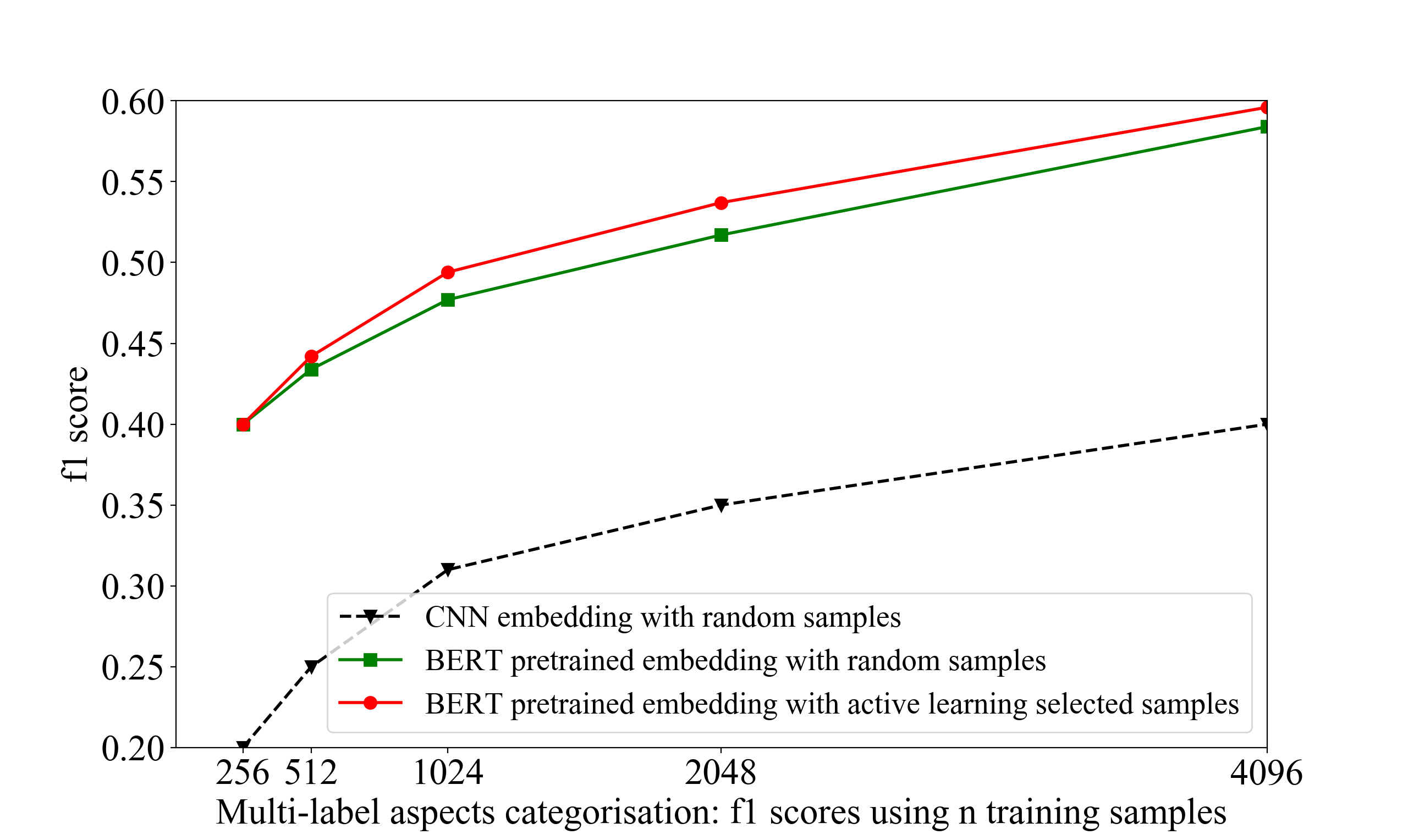}
	\caption{The evaluations on the multi-label classification tasks: aspects categorisation (13 classes) and sentiment analysis (2 classes), the reported f1 score is measured with incremental training sizes.} 
	\label{fig:results}
\end{figure}

We report micro f1 scores under different training sizes in Fig.\;\ref{fig:results}.  The micro f1 scores is calculated in a multi-label situation using micro precision and micro recall summed over all classes, shown as follows:
 
\begin{equation}
	\small
\begin{split}
  \text{micro precision} = & \quad \frac{\sum_{i = 1}^{C} \text{true positive}}{\sum_{i = 1}^{C} \text{true positive} + \sum_{i = 1}^{C} \text{false positive}} \\
  \text{micro recall} = & \quad  \frac{\sum_{i = 1}^{C} \text{true positive}}{\sum_{i = 1}^{C} \text{true positive} + \sum_{i = 1}^{C} \text{false negative}} 
\end{split}
\end{equation}
\begin{equation}
	\small
\text{micro f1 score}  =   \quad  2 \times \frac{\text{micro precision} \times  \text{micro recall}}{\text{micro precision} + \text{micro recall}} 
\end{equation}
Such choice of measures follows the standard in the literature for multi-label classification \cite{pontiki2016semeval}, especially in the case of highly unbalanced classes. We report the comparison in Fig.\;\ref{fig:results} using the average micro f1 score over three times of different independent experiments. 

As we can see in Fig.\;\ref{fig:results}, in case of sentiment classification, the network with CNN embedding yield the worst performance in all different training sample sizes, due to the lack of effectiveness of language representations by training the embedding with largely insufficient data. 

Such a gap between the self-trained embedding and the pretrained embedding with BERT is even large when facing more challenging multi-label aspects categorisation tasks. 

When comparing the data selection framework, we can see in both tasks that the active learning selection performs better. In other words, in order to achieve the same performance, an active learning framework needs much fewer data.  

Please note that the reported f1 scores with active learning are based on one straightforward selection strategy as in Eq.\;\eqref{eq:certain}. By analyzing the empirical results in the active learning literature \cite{sener2017active},  we firmly believe that the reported results can be further improved when using some of the sophisticated selection strategies.

\section{Conclusion}

In this paper, we introduce two  strategies for boosting the performance in real-world applications for natural language processing: 1. pretrained language model that allows extracting essential features of texts without the extra effort of collecting a large amount of training data; 2. active learning strategy which can smartly select the samples that need to be labeled. By comparing the performance with basic recurrent neural networks and the ones combined with pretrained embedding model and active learning framework, we observe a significant improvement. Such a combined approach can achieve the same accuracy by using a significantly smaller amount of labeled data, thus provide cost-effective solutions for the company self-promoted natural language processing projects. In the future, we would like to investigate on more sophisticate active learning strategies, in order to further improve the results with a constant number of training size.  

\subsubsection*{Acknowledgment}

The authors would like to thank Isabelle DUPUIS, Nathalie CHANSON, Axelle LETERTRE, Isabelle ROMANO from ``Voice of the Customer" at GMF ASSURANCES for their expertises in customer review analysis and providing the labeled data used in this paper.


\bibliographystyle{IEEEbib}
\bibliography{refs}

\end{document}